\documentclass{article}
\usepackage{arxiv}
\usepackage{hyperref}
\usepackage[linesnumbered,ruled,vlined]{algorithm2e}
\usepackage{array}
\usepackage[caption=false,font=normalsize,labelfont=sf,textfont=sf]{subfig}
\usepackage{textcomp}
\usepackage{stfloats}
\usepackage{url}
\usepackage{verbatim}
\usepackage{graphicx}
\usepackage{rotating}
\usepackage{balance}
\usepackage{multirow}
\usepackage{rotating}
\usepackage{tabularx}
\usepackage{booktabs}
\usepackage{amsmath}

\usepackage{xcolor}
\usepackage[normalem]{ulem}
\useunder{\uline}{\ul}{}

\title{KAN KAN Buff Signed Graph Neural Networks?}

\author{Muhieddine Shebaro and Jelena Te\v{s}i\'{c}}

\begin{document}

\maketitle
\begin{abstract} Graph Representation Learning aims to create effective embeddings for nodes and edges that encapsulate their features and relationships. Graph Neural Networks (GNNs) leverage neural networks to model complex graph structures. Recently, the Kolmogorov-Arnold Neural Network (KAN) has emerged as a promising alternative to the traditional Multilayer Perceptron (MLP), offering improved accuracy and interpretability with fewer parameters. In this paper, we propose the integration of KANs into Signed Graph Convolutional Networks (SGCNs), leading to the development of KAN-enhanced SGCNs (KASGCN). We evaluate KASGCN on tasks such as signed community detection and link sign prediction to improve embedding quality in signed networks. Our experimental results indicate that KASGCN exhibits competitive or comparable performance to standard SGCNs across the tasks evaluated, with performance variability depending on the specific characteristics of the signed graph and the choice of parameter settings. These findings suggest that KASGCNs hold promise for enhancing signed graph analysis with context-dependent effectiveness.
\end{abstract}
Kolmogorov-Arnold Neural Networks, Signed Network, Graph Representation Learning, Graph Neural Networks

Graph Representation Learning is creating embeddings for nodes and edges based on the features and connections in a graph. These embeddings capture the graph's local and global features and are useful for node classification and link prediction tasks. One of the most well-known methods for learning graph embeddings is through Graph Neural Networks (GNNs) \cite{emb}. GNNs have proven to understand complex graph relationships and structures effectively. Graph neural networks gather information from neighboring nodes, gradually refining the embeddings through multiple layers \cite{appsocial,9903566}. The Kolmogorov-Arnold Neural Network (KAN) has emerged recently as an alternative to the traditional Multilayer Perceptron (MLP) architecture \cite{liu2024kankolmogorovarnoldnetworks}.
In contrast to traditional multilayer perceptrons (MLPs), KAN utilizes learnable univariate functions instead of fixed activation functions. KANs have outperformed MLPs for the numerical analysis and partial differential equation solving \cite{ji2024comprehensivesurveykolmogorovarnold} as the learnable activation functions on edges show more flexibility in learning complex models, and KAN can also handle high-dimensional data. The fusion of KANs and graph neural networks in several ways has improved the performance and the quality of the GNN embeddings \cite{decarlo2024kolmogorovarnoldgraphneuralnetworks,kiamari2024gkangraphkolmogorovarnoldnetworks,bresson2024kagnnskolmogorovarnoldnetworksmeet} in the node classification, link prediction, and graph classification tasks.

This paper introduces the first known integration of the Kolmogorov-Arnold Neural Network (KAN) and its variants into signed Graph Neural Networks (GNNs), evaluating their impact on tasks such as signed clustering and link sign prediction. Additionally, we investigate how KANs can enhance the quality of embeddings generated by signed GNNs. Signed GNNs are particularly suited for graphs where edges represent positive or negative interactions, capturing the dynamics of various relationships. The Signed Graph Convolutional Networks (SGCNs) method, proposed by Derr et al., adapts Graph Convolutional Networks to signed networks by leveraging balance theory \cite{derr2018signedgraphconvolutionalnetwork}. Balance theory, which addresses the dynamics of attitudes within networks, has applications in edge sentiment prediction and anomaly detection. In this context, a signed network is considered strongly balanced if every fundamental cycle contains an even number of negative edges.

Signed Graph Convolutional Neural Network (SGCN) is the baseline and the most robust variant of signed GNNs in the field, as it adapts the Graph Convolutional Neural Network for signed networks based on the balance theory \cite{derr2018signedgraphconvolutionalnetwork}. Balance theory is pivotal in explaining attitudes' evolution within signed graph networks. Heider formalized the balance theory in \cite{1958Abelson}, and Harary introduced the mathematical formulation and the $k$-way balancing~\cite{Har2, Harary1968}. The balance theory has been widely applied in various domains, including edge sentiment prediction, content and product recommendations, and anomaly detection~\cite{derr2020link,garimella2021political,interian2022network,amelkin2019fighting}. In the context of signed networks, a network is considered strongly balanced if every fundamental cycle consists of an even number of negative edges. The Signed Graph Convolutional Network (SGCN) incorporates the balance theory by maintaining two distinct representations at each layer: one for the suggested friends, where the path connecting the node contains an even number of negative links, and another for the suggested enemies, where the path includes an odd number of negative links.

\begin{figure*}[!ht]
   \centering
    \includegraphics[width=\linewidth]{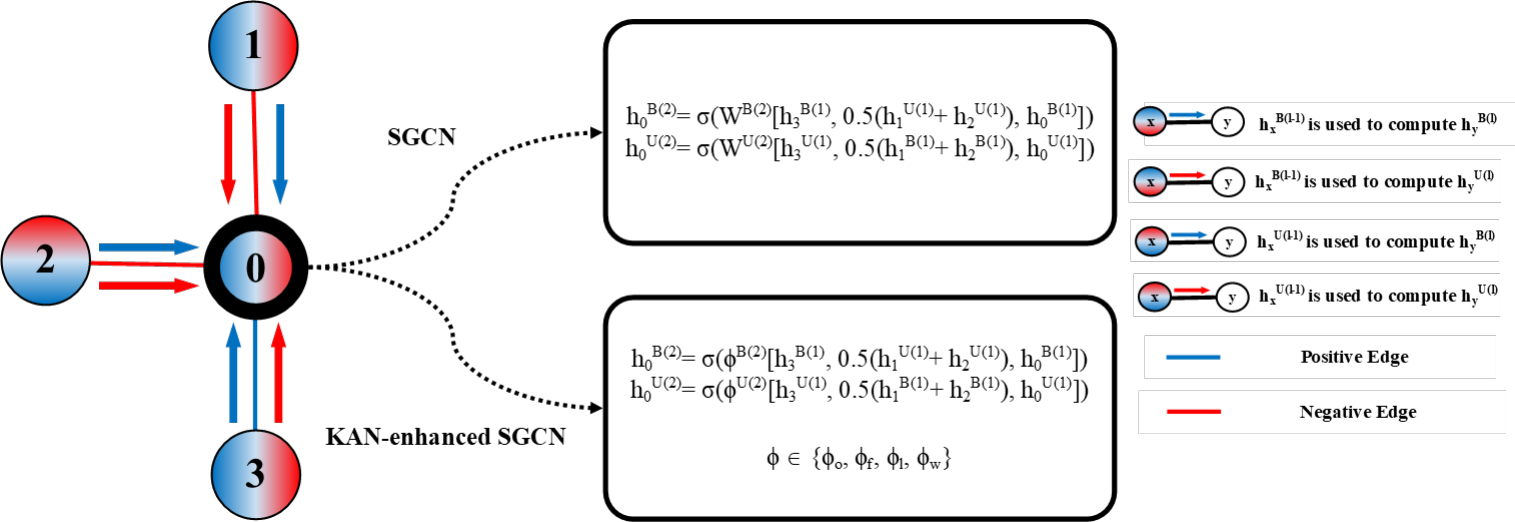}
    \caption{Illustration of SGCN and KAN-enhanced SGCN of a signed network of 4 nodes and three edges.} \label{fig-kan_illust}
\end{figure*}
    
This research project is the first to integrate KAN (with its variants) into signed graph neural networks and empirically evaluate its impact on performance on two downstream tasks: signed clustering and link sign prediction for signed networks. Next, the project outlines KANs' potential to increase the expressive power of the embeddings generated by signed graph neural networks. Section~\ref{ssec-related} discusses the related work regarding KAN integration into several architectures, their uses, and the background on KAN and SGCN. In Section~\ref{sec-metho}, we present the methodology of our work and pose research questions for this study. In Section~\ref{sec-exp}, we empirically evaluate the KAN-integrated signed GNN and its variants in several tasks, including efficiency examination and embedding comparison. We structure this section by presenting the hypothesis, results, and observations. In Section~\ref{sec-summ}, we summarize our findings.

\subsection{Related Work}
\label{ssec-related}

Cheon et al. proposed a method that combines Kolmogorov-Arnold Networks (KANs) with pre-trained Convolutional Neural Networks (CNNs), specifically VGG16 and MobileNetV2 models, for scene classification in satellite imagery \cite{cheon2024kolmogorovarnoldnetworksatelliteimage}. The proposed approach demonstrated high accuracy, requiring fewer epochs and parameters than traditional Multilayer Perceptrons (MLPs) that do not integrate KANs \cite{cheon2024kolmogorovarnoldnetworksatelliteimage}. KANs have been explored for data representation using autoencoders and compared against conventional Convolutional Neural Networks (CNNs) on datasets such as MNIST, SVHN, and CIFAR-10. And it is shown that KAN-based autoencoders deliver competitive reconstruction accuracy \cite{moradi2024kolmogorovarnoldnetworkautoencoders}. Dong et al. evaluated the performance of KANs on time series data and found that KANs can match or even surpass the performance of MLP across 128 different time series datasets \cite{dong2024kolmogorovarnoldnetworkskantime}. Lu et al. assessed using KANs for fraud detection and concluded that their performance varies depending on the context \cite{lu2024kolmogorovarnoldnetworksfraud}. The study in \cite{anonymous2024graphkan} presented a proof of concept for employing KANs in graphs to predict molecules' binding affinity to protein targets. The model in this study did not attain state-of-the-art performance, but it emphasized its promising method in computational drug discovery. For the Graph Neural Network integration, Bresson et al. integrated KAN layers into unsigned graph neural networks (GNNs) for the node and graph classification tasks \cite{bresson2024kagnnskolmogorovarnoldnetworksmeet}, and Li et al. integrated Fourier series-based KANs to optimize GNNs further for the molecular property prediction. \cite{li2024kagnnkolmogorovarnoldgraphneural}. 


\section{Introducing the Kolmogorov Arnold Signed Graph Convolutional Networks}
\label{sec-KASGCN}

\subsection{Kolmogorov-Arnold Networks}
\label{ssec-kan}

The Kolmogorov-Arnold representation theorem states that any multivariate function can be formulated as a combination of continuous univariate functions \cite{Kolmogorov1956}, as defined in Eq.~\ref{eq-kan1}. \begin{equation}
    f(x_1, x_2, \ldots, x_n) = \sum_{i=1}^{2d+1} \alpha_i \left( \sum_{j=1}^{d} \phi_{ij}(x_j) \right) \label{eq-kan1}
\end{equation}

\noindent The parameters $\alpha$ and $\phi$ are the univariate functions and $d$ is the dimension of the input. The output is an aggregated output of each input after it passes through an univariate and non-linear spline function. The main hyperparameters for the spline function are the spline order (degree of B-splines) and the grid size (number of intervals to approximate the real function).

Kolmogorov-Arnold Networks (KANs) demonstrate greater expressiveness than standard multilayer perceptrons (MLPs) and superior performance with significantly fewer parameters \cite{liu2024kankolmogorovarnoldnetworks}. Where MLPs adjust the weights globally based on the training data, KANs spline control points impact only local regions, thereby preserving more information \cite{controlpoint}. Next, $\phi(x)$ from Eq.~\ref{eq-kan1} is defined in Eq.~\ref{eq-phi}. 

\begin{equation}
    \phi(x) = w_b*\text{SiLU}(x) + w_s * \text{spline}(x). \label{eq-phi}  
\end{equation}

\noindent The $\text{SiLU}(x)$ is the SiLU activation function $x/(1+e)$ and $\text{spline}(x)$ is defined as the is a linear combination of B-splines, $\text{spline}(x) = \sum_i c_i B_i(x)$ where $c_i, w_b$, and $w_s$ are trainable and control the magnitude of the activation function. $w_s = 1$ and $w_b$ are initialized using Xavier initialization.

\subsection{Signed Graph Convolutional Neural Networks}
\label{ssec-sgcn}
Derr et al. apply balance theory to effectively aggregate features from a node's neighbors in signed networks \cite{derr2018signedgraphconvolutionalnetwork}. Each node has a positive embedding and a negative embedding representation. During the message-passing process, the positive embedding updates by aggregating the positive embeddings from its positive neighbors. The negative embeddings update from its negative neighbors, concatenating the resulting aggregate to the node's positive representation. By aggregating the negative embeddings from positive neighbors and the positive embeddings from negative neighbors concatenate with the node's negative representation.

Conversely, the negative embedding is updated. The final node embedding is obtained by concatenating both the positive and negative embeddings. The signed graph convolutional network (GCN) formulation is described as follows.

\noindent For $l = 1:$
\[
    h_i^{B(1)}=\sigma(W^{B(1)}[\sum_{j \in N_i^+}{\frac{h_j^{0}}{|N_i^+|}},h_i^{0}])
\]
\[
    h_i^{U(1)}=\sigma(W^{U(1)}[\sum_{k \in N_i^+}{\frac{h_k^{0}}{|N_i^+|}},h_i^{0}])
\]

\noindent For $l > 1:$
\[
    h_i^{B(l)}=\sigma(W^{B(l)}[\sum_{j \in N_i^+}{\frac{h_j^{B(l-1)}}{|N_i^+|}},\sum_{k \in N_i^-}{\frac{h_k^{U(l-1)}}{|N_i^-|}},h_i^{B(l-1)}])
\]
\[
    h_i^{U(l)}=\sigma(W^{B(l)}[\sum_{j \in N_i^+}{\frac{h_j^{U(l-1)}}{|N_i^+|}},\sum_{k \in N_i^-}{\frac{h_k^{B(l-1)}}{|N_i^-|}},h_i^{U(l-1)}])
\]

\noindent The number of layers is $l$, $h_i^{B(l)}$ and $h_i^{U(l)}$ are the positive and negative representations of node $i$  in layer $l$ respectively, $\sigma$ is an activation function, $W^{B(l)}$ and $W^{U(l)}$ are the weight matrices for updating the positive and negative representations respectively, $N_i^+$ and $N_i^-$ are the sets of positive and negative neighbors of $i$ respectively. 

\subsection{Outline of the KASGCN Method}
\label{sec-metho}

In this section, we introduce the Kolmogorov-Arnold Signed Graph Convolutional Network (KASGCN) method to integrate the Kolmogorov-Arnold layer (KAN) in the signed graph convolutional neural network (SGCN) by replacing the weight matrix $W$ with the KAN layer in the formulation. The KAN layers have trainable parameters and introduce non-linearity, superseding the role of the weight matrix illustrated for the unsigned example in \cite{bresson2024kagnnskolmogorovarnoldnetworksmeet}. We formalize the Kolmogorov-Arnold signed graph convolutional network (KASGCN) method as follows: 

\noindent For $l = 1:$

\[
    h_i^{B(1)}=\sigma(\phi^{B(1)}([\sum_{j \in N_i^+}{\frac{h_j^{0}}{|N_i^+|}},h_i^{0}]))
\]
\[
    h_i^{U(1)}=\sigma(\phi^{U(1)}([\sum_{k \in N_i^+}{\frac{h_k^{0}}{|N_i^+|}},h_i^{0}]))
\]

\noindent For $l > 1:$
\[
    h_i^{B(l)}=\sigma(\phi^{B(l)}([\sum_{j \in N_i^+}{\frac{h_j^{B(l-1)}}{|N_i^+|}},\sum_{k \in N_i^-}{\frac{h_k^{U(l-1)}}{|N_i^-|}},h_i^{B(l-1)}]))
\]
\[
    h_i^{U(l)}=\sigma(\phi^{U(l)}([\sum_{j \in N_i^+}{\frac{h_j^{U(l-1)}}{|N_i^+|}},\sum_{k \in N_i^-}{\frac{h_k^{B(l-1)}}{|N_i^-|}},h_i^{U(l-1)}]))
\]

\noindent the  $\phi^{B(l)}$ and $\phi^{U(l)}$ here represent the KAN layer used to update the positive and negative representations, respectively. This layer can be any state-of-the-art variant such as OriginalKAN $\phi_o$ \cite{liu2024kankolmogorovarnoldnetworks}, FourierKAN $\phi_f$ \cite{xu2024fourierkangcffourierkolmogorovarnoldnetwork}, LaplaceKAN $\phi_l$ \cite{Toy-KAN}, and WaveletKAN $\phi_w$ \cite{bozorgasl2024wavkanwaveletkolmogorovarnoldnetworks}. We later experimented with these variants to determine how each performed when integrated into SGCN. As an illustration, Figure~\ref{fig-kan_illust} depicts both the SGCN and the KAN-enhanced SGCN starting from $l = 1$. In addition, we pose the following research questions, and we answer them in Section~\ref{sec-exp}:

\begin{enumerate}
    \item Does incorporating the KAN layer into the signed graph convolutional network improve the clustering quality when using Kmeans++ on the embeddings?
    \item Do the embeddings generated from integrating the KAN layer into SGCN increase the predictive power of logistic regression for sign classification?
    \item How efficient is KASGCN compared to SGCN?
    \item How similar do SGCN and KASGCN produce the embeddings?
    \item Does integrating other KAN variants into SGCN perform just as well as integrating the original KAN layer?
\end{enumerate}

\section{Proof Of Concept Setup}
\label{sec-setup}
\subsection{Setup}
The operating system used for the experiments is Linux Ubuntu 20.04.3, running on the 11th Gen Intel(R) Core(TM) i9-11900K @ 3.50GHz with 16 physical cores. It is one socket, with two threads per core and eight cores per socket. The architecture is X86\_x64. The GPU is Nvidia GeForce RTX 3070 and has 8GB of memory. Its driver version is 495.29.05, and the CUDA version is 11.5. The cache configuration is L1d : 384 KiB, L1i : 256 KiB, L2 : 4 MiB, L3 : 16 MiB. The CPU op is 32-bit and 64-bit.
\subsection{Parameters}
 Each KAN layer consists of a grid of splines. The degree of each spline is the \emph{spline order}, and the number of splines for each function is known as grid size. For the implementation of B-splines and other variants of KAN, we rely on the publicly available implementations in \cite{Blealtan2024} and \cite{Toy-KAN}, respectively. The parameters used in the experiments are: layers = [32,32], weight\_decay = $10^{-5}$, learning\_rate = 0.001, lamb = 1.0, epochs = 1000, seed = 42, reduction\_iterations = 10, reduction\_dimensions = 15, spectral\_features = True, norm = True, norm\_embed = True, grid\_size = 5,  spline\_order = 3, scale\_noise = 0.1, scale\_base = 1.0, scale\_spline = 1.0, base\_activation = torch.nn.SiLU, grid\_eps = 0.02, grid\_range= [-1, 1], Kmeans++ is used for clustering and with default parameters. Each experiment is repeated 10 times and averaged. The test size split for each positive and negative for link sign prediction (using multinomial logistic regression) is 0.2. t-SNE parameters are random\_state = 0, n\_iter = 1000, metric = 'cosine'. Pre-processing of the signed network includes Removing duplicates and keeping a first edge, treating neutral edges as positive, dropping self-loops, reindex nodes from 0 to $n-1$ where $n$ is the number of nodes in the signed network. The embedding passes through the KAN layer and the tanh activation function. 
 
\subsection{Evaluation Metrics}
We use two metrics to gauge the quality of our clustering assignments: fraction of positive edges within clusters and fraction of negative edges between clusters). They are the following:

\begin{equation}
 pos_{in}= \frac{pos_{within}}{pos_{between}+pos_{within}}
  \label{eq-PosIn} 
\end{equation}
\begin{equation}
neg_{out}= \frac{neg_{between}}{neg_{within}+neg_{between}}
  \label{eq-NegOut} 
\end{equation}

\noindent where $pos_{within}$ and $pos_{between}$ are the number of positive edges within and between clusters, respectively. $neg_{within}$ and $neg_{between}$ are the negative edges within and between clusters. To measure the overall clustering quality $Q$, we add both to obtain:
\begin{equation}
    Q =  pos_{in} + neg_{out}
    \label{eq-Q}
\end{equation}

Moreover, we rely on the AUC score that summarizes the GNN's effectiveness across all possible classification thresholds to evaluate the link sign prediction performance. In addition, we use the F1 score for this as well, which is as follows:

\begin{equation}
\text{F1} = 2 \cdot \frac{\text{Precision} \cdot \text{Recall}}{\text{Precision} + \text{Recall}}
\end{equation}

\noindent where $Precision$ is the ratio of true positive instances to the sum of true positive and false positive instances. And $Recall$ is the ratio of true positive instances to the sum of true positive and false negative instances. We use seconds to measure and compare the training time of the graph neural network (GNN) models. The average cosine similarity is employed to measure the  similarity between the embeddings generated by two different GNNs for each node in the signed network, and the formula is:

\begin{equation}
\text{Avg Cosine Sim} = \frac{1}{N} \sum_{i \in N} \frac{\mathbf{e}_i^A \cdot \mathbf{e}_i^B}{\|\mathbf{e}_i^A\| \|\mathbf{e}_i^B\|}
\end{equation}

\noindent where $N$ is the set of all nodes in the network, $\mathbf{e}_i^A$ and $\mathbf{e}_i^B$ are the embeddings of node $i$ generated from GNN $A$ and GNN $B$ respectively.

\subsection{Signed Graphs}
Table~\ref{tab-KonectScale} describes the Konect and other signed graphs and their characteristics \cite{konect}. \em{BitcoinAlpha} is a user-user trust/distrust network from the Bitcoin Alpha platform for trading bitcoins. \em{BitcoinOTC} is a user-user trust/distrust network from the Bitcoin OTC platform for trading Bitcoins. \emph{WikiRFA} describes voting information for electing Wikipedia managers \cite{he2022sssnetsemisupervisedsignednetwork}. \em{WikiElec} is the network of users from the English Wikipedia that voted for and against each other in admin elections. In the \emph{Chess} network, each vertex is a chess player, and a directed edge represents a game with the white player having an outgoing edge and the black player having an ingoing edge. The weight of the edge represents the outcome. \em{Congress} is a signed network where vertices are politicians speaking in the United States Congress, and a directed edge denotes that a speaker mentions another speaker. \emph{PPI} models the protein-protein interaction network \cite{he2022sssnetsemisupervisedsignednetwork}. Preprocessing removes duplicates (keep first), treats neutral edges as positive, removes self-loops, and reindexes nodes from 0 to the maximum number of nodes in the signed graph.

\begin{table*}[!ht]
\setlength\tabcolsep{1pt}
\centering
\caption{Konect plus WikiRFA and PPI properties. Vertices, edges, and cycles represent the signed graph, not just the largest connected component (LCC). The rest of the metrics correspond to the LCC.}
\label{tab-KonectScale}
\begin{tabular}{cccccccccc}
\toprule
\textbf{Graph} & \textbf{Vertices}&\textbf{Edges}& \textbf{Cycles} &\textbf{Density} & \textbf{Triads} & \textbf{Avg Deg.} & \textbf{Median deg.} & \textbf{Max deg.} &\textbf{\% of $e^-$} \\
\midrule
BitcoinAlpha &3,775 &14,120 &10,346 & $<$ 0.01 & 22,153 & 7.48 & 2& 511& 8.39 \\
BitcoinOTC & 5,875 & 21,489 & 15,615 & 0.01 & 33,493 & 7.31 &2 &795 & 13.57 \\
WikiRFA & 7,634 & 167,936 & 160,303  &  $<$ 0.01 & 1,240,033 & 43.99 & 13 & 1,223 & 22.98 \\
WikiElec &7,066& 100,667 & 93,602 & $<$ 0.01 & 607,279 & 28.49 &4 &1,065 & 21.94 \\
Chess & 7,115 & 55,779 & 48,665 &$<$ 0.01 & 108,584 & 15.67 & 7 &181 & 24.15\\
Congress & 219 & 521 &303 & 0.021 & 212 & 4.71 &3 &33 & 20.34 \\
PPI & 3,058 & 11,860 & 8,803 & $<$ 0.01 & 3,837 & 3.87 &2 &55 & 32.5 \\
\bottomrule
\end{tabular}
\end{table*}

\section{Experimental Results}
\label{sec-exp}
\subsection{Community Detection}
\begin{itemize}
    \item \textbf{Hypothesis:} KASGCN learns representations that enhance the clustering of vertices by minimizing the negative edges between different clusters and maximizing the number of positive edges within the same cluster.
    \item \textbf{Results:} The clustering results are summarized in Table~\ref{tab-com1}, Table~\ref{tab-com2}, and Table~\ref{tab-com3} for $K = 5$, $K = 10$, $K = 15$ respectively where $K$ is the predefined number of clusters parameter as input for the Kmeans++ algorithm.
    
\begin{table*}[!ht]
\centering
\caption{Average $pos\_{in}$ and $neg\_{out}$ over 10 runs for SGCN and KASGCN node embeddings using Kmeans++ clustering ($K = 5$). Gain is computed using $\frac{100 \cdot(Q_{KASGCN}-Q_{SGCN})}{Q_{SGCN}}$.}\label{tab-com1}
    \begin{tabular}{l|cc|cc|r} \toprule
        \textbf{GNN} & \multicolumn{2}{c|}{\textbf{SGCN}}  & \multicolumn{2}{c|}{\textbf{KASGCN}} & \textbf{~} \\ 
        \textbf{Quality} & $pos$ & $neg$ &  $pos$ & $neg$ & Gain \\ \midrule
BitcoinAlpha & 0.388 $\pm$ 0.0003 & 0.8015 $\pm$ 0.003 & 0.4093 $\pm$ 0.011 & 0.659 $\pm$ 0.05 & -10.18\% \\ 
BitcoinOTC & 0.513 $\pm$ 0.011 & 0.820 $\pm$ 0.054 & 0.523 $\pm$ 0.00033 & 0.707 $\pm$ 0.001 & -7.70\% \\ 
WikiElec & 0.507 $\pm$ 0.0002 & 0.773 $\pm$ 0.0004 & 0.5221 $\pm$ 0.004 & 0.784 $\pm$ 0.0009 & 2.03\% \\  
WikiRFA & 0.410 $\pm$ 0.0025 & 0.843 $\pm$ 0.0021 & 0.391 $\pm$ 0.0010 & 0.795 $\pm$ 0.0006 & -5.34\% \\ 
Chess & 0.434 $\pm$ 0.002 & 0.641 $\pm$ 0.014 & 0.409 $\pm$ 0.015 & 0.619 $\pm$ 0.034 & -4.27\% \\ 
Congress & 0.616 $\pm$ 0.028 & 0.961 $\pm$ 0.009 & 0.566 $\pm$ 0.029 & 0.9584 $\pm$ 0.017 & -3.33\% \\ 
PPI & 0.495 $\pm$ 0.001 & 0.709 $\pm$ 0.010 & 0.385 $\pm$ 0.084 & 0.87 $\pm$ 0.0109 & 4.23\% \\ \bottomrule
    \end{tabular}
\end{table*}

\begin{table*}[!ht]
    \centering
    \caption{Average $pos\_{in}$ and $neg\_{out}$ over 10 runs for SGCN and KASGCN node embeddings using Kmeans++ clustering ($K = 10$). Gain is computed using $\frac{100 \cdot(Q_{KASGCN}-Q_{SGCN})}{Q_{SGCN}}$.}\label{tab-com2}
    \begin{tabular}{l|cc|cc|r}
\toprule
        \textbf{GNN} & \multicolumn{2}{c|}{\textbf{SGCN}}  & \multicolumn{2}{c|}{\textbf{KASGCN}} & \textbf{~} \\ 
        \textbf{Quality} & $pos$ & $neg$ &  $pos$ & $neg$ & Gain \\ \midrule
        BitcoinAlpha & 0.275 $\pm$ 0.023 & 0.911 $\pm$ 0.0040 & 0.299 $\pm$ 0.006 & 0.877 $\pm$ 0.012 & -0.84\% \\ 
        BitcoinOTC & 0.378 $\pm$ 0.020 & 0.929 $\pm$ 0.0045 & 0.421 $\pm$ 0.005 & 0.864 $\pm$ 0.051 & -1.60\% \\ \
        WikiElec & 0.3260 $\pm$ 0.005 & 0.8956 $\pm$ 0.003 & 0.283 $\pm$ 0.020 & 0.909 $\pm$ 0.0022 & -2.40\% \\ 
        WikiRFA & 0.2637 $\pm$ 0.004 & 0.910 $\pm$ 0.0013 & 0.270 $\pm$ 0.033 & 0.917 $\pm$ 0.04 & 1.13\% \\ 
        Chess & 0.254 $\pm$ 0.026 & 0.798 $\pm$ 0.038 & 0.22 $\pm$ 0.0054 & 0.888 $\pm$ 0.0022 & 5.30\% \\ 
        Congress & 0.497 $\pm$ 0.023 & 0.996 $\pm$ 0.0048 & 0.472 $\pm$ 0.050 & 0.992 $\pm$ 0.0074 & -1.80\% \\ 
        PPI & 0.422 $\pm$ 0.0366 & 0.784 $\pm$ 0.011 & 0.364 $\pm$ 0.037 & 0.936 $\pm$ 0.0093 & 7.79\% \\ \bottomrule
    \end{tabular}
\end{table*}

\begin{table*}[!ht]
    \centering
    \caption{Average $pos\_{in}$ and $neg\_{out}$ over 10 runs for SGCN and KASGCN node embeddings using Kmeans++ clustering ($K = 15$). Gain is computed using $\frac{100 \cdot(Q_{KASGCN}-Q_{SGCN})}{Q_{SGCN}}$.}\label{tab-com3}
       \begin{tabular}{l|cc|cc|r} \toprule 
        \textbf{GNN} & \multicolumn{2}{c|}{\textbf{SGCN}}  & \multicolumn{2}{c|}{\textbf{KASGCN}} & \textbf{~} \\
        \textbf{Quality} & $pos$ & $neg$ &  $pos$ & $neg$ & Gain \\ \midrule
BitcoinAlpha & 0.2266 $\pm$ 0.010 & 0.943 $\pm$ 0.0070 & 0.245 $\pm$ 0.0074 & 0.921 $\pm$ 0.006 & -0.30\% \\
BitcoinOTC & 0.335 $\pm$ 0.015 & 0.956 $\pm$ 0.005 & 0.380 $\pm$ 0.17 & 0.93 $\pm$ 0.014 & 1.47\% \\
WikiElec & 0.249 $\pm$ 0.067 & 0.935 $\pm$ 0.002 & 0.227 $\pm$ 0.0090 & 0.942 $\pm$ 0.04 & -1.20\% \\
WikiRFA & 0.197 $\pm$ 0.0072 & 0.9407 $\pm$ 0.037 & 0.203 $\pm$ 0.0051 & 0.9470 $\pm$ 0.0050 & 1.08\% \\ 
Chess & 0.195 $\pm$ 0.003 & 0.853 $\pm$ 0.0033 & 0.181 $\pm$ 0.0024 & 0.922 $\pm$ 0.0020 & 5.24\% \\ 
Congress & 0.438 $\pm$ 0.027 & 0.997 $\pm$ 0.0045 & 0.392 $\pm$ 0.034 & 0.996 $\pm$ 0.0048 & -3.27\% \\ 
PPI & 0.333 $\pm$ 0.01 & 0.833 $\pm$ 0.0064 & 0.309 $\pm$ 0.031 & 0.959 $\pm$ 0.0049 & 8.74 \\ \bottomrule
    \end{tabular}
\end{table*}

    \item \textbf{Observations:} Improvement in the clustering quality highly depends on the choice of K. For relatively higher values of K, Kmeans++ using embeddings from KASGCN seems to do slightly better. In comparison, for relatively lower values of K, Kmeans++ using embeddings from SGCN appears to do slightly better. Overall, the difference in the clustering quality between SGCN and KASGCN is negligible or meager except in PPI, where KASGCN outperforms SGCN regardless of K.
\end{itemize}

 \begin{table*}[!ht]
    \centering
    \caption{Average AUC and F1 score over 10 runs for SGCN and KASGCN node embeddings using Multinomial Logistic Regression with random 0.2 test size split.} \label{tab-link}
   \begin{tabular}{l|cc|cc|rr} \toprule
\textbf{GNN} & \multicolumn{2}{c|}{\textbf{SGCN}}& \multicolumn{2}{c|}{\textbf{KASGCN} }& \multicolumn{2}{c}{\textbf{Gain}} \\
\textbf{Metric} & AUC & F1 & AUC & F1 & AUC & F1 \\ \midrule
BitcoinAlpha & 0.783 $\pm$ 0.014 & 0.721 $\pm$ 0.042 & 0.799 $\pm$ 0.022 & 0.690 $\pm$ 0.094 & 2.04\% & -4.29\% \\
BitcoinOTC & 0.840 $\pm$ 0.005 & 0.792 $\pm$ 0.014 & 0.859 $\pm$ 0.0075 & 0.761 $\pm$ 0.068 & 2.26\% & -3.91\% \\
WikiElec & 0.809 $\pm$ 0.0044 & 0.781 $\pm$ 0.0087 & 0.837 $\pm$ 0.0045 & 0.787 $\pm$ 0.032 & 3.46\% & 0.76\% \\ 
WikiRFA & 0.820 $\pm$ 0.0031 & 0.7623 $\pm$ 0.014 & 0.830 $\pm$ 0.0037 & 0.759 $\pm$ 0.017 & 1.21\% & -0.434 \\
Chess & 0.542 $\pm$ 0.005 & 0.612 $\pm$ 0.017 & 0.548 $\pm$ 0.0064 & 0.625 $\pm$ 0.041 & 1.10\% & 2.12\% \\
Congress & 0.570 $\pm$ 0.055 & 0.63 $\pm$ 0.058 & 0.499 $\pm$ 0.046 & 0.602 $\pm$ 0.077 & -12.29\% & -4.44\% \\
PPI & 0.679 $\pm$ 0.015 & 0.614 $\pm$ 0.067 & 0.692 $\pm$ 0.015 & 0.654 $\pm$ 0.037 & 1.91\% & 6.51\% \\ \bottomrule
    \end{tabular}
\end{table*}

\subsection{Link Sign Prediction}
\begin{itemize}
    \item \textbf{Hypothesis:} The expressiveness of KASGCN's embeddings allows logistic regression to better predict the sign of the edge than that of SGCN.
    \item \textbf{Results:}
    The link sign prediction results are presented in Table~\ref{tab-link}, where a Multinomial Logistic Regression is trained on the embeddings of both SGCN and KASGCN with random 0.2 test size splits over 10 runs.
    \item \textbf{Observations:} Logistic Regression with embeddings from KASGCN shows consistent improvement in AUC across most datasets, indicating better overall performance in distinguishing between classes. The highest AUC improvement is observed in the WikiElec dataset, with a 3.46\% increase. The F1 score improvements are mixed. While there are gains in datasets like WikiElec (0.76\%) and Chess (2.12\%), there are notable declines in datasets like BitcoinAlpha (-4.29\%) and BitcoinOTC (-3.91\%). While Logistic Regression with embeddings from KASGCN generally enhances AUC, the impact on F1 scores is graph-dependent.
\end{itemize}

\subsection{Efficiency Examination}
\begin{itemize}
    \item \textbf{Hypothesis:} Incorporating the KAN layer into the Signed Graph Neural Network increases the model's complexity, leading to longer execution times for training embeddings.
    \item \textbf{Results:} We visualize the efficiency of training SGCN and KASGCN embeddings in Figure~\ref{fig-efficiency} with varying numbers of aggregator layers.

    \begin{figure*}[!ht]
        \centering
        \includegraphics[width=\textwidth]{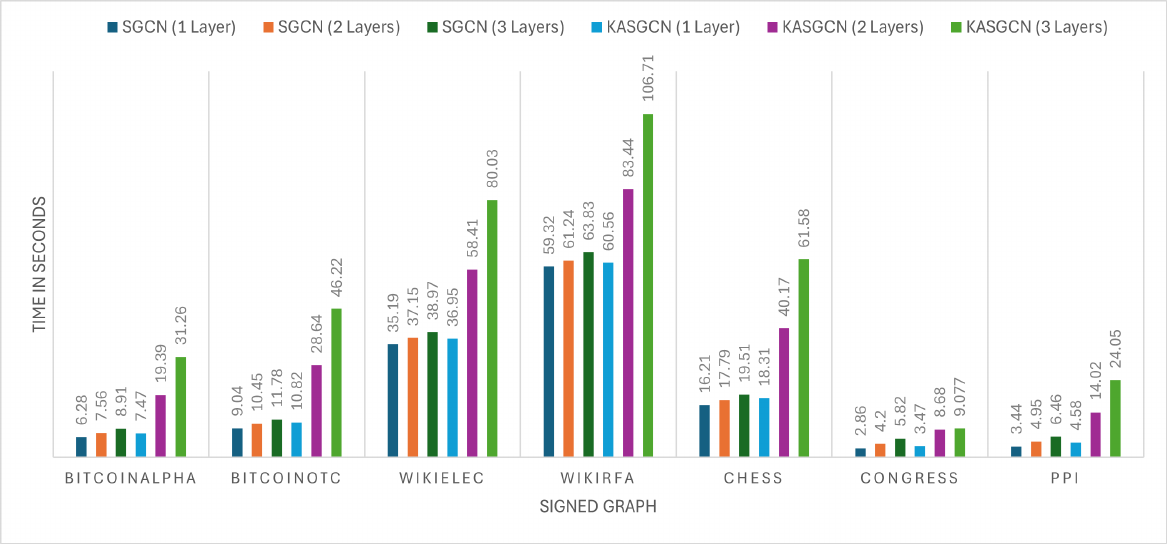}
        \caption{The embedding generation time of SGCN and KASGCN with varying aggregator layers.}
        \label{fig-efficiency}
    \end{figure*}
    \item \textbf{Observations:} As expected, due to the spline function, KAGCN takes a significantly longer time to train and generate embeddings. Increasing the aggregation layers increases the execution time for the KASGCN but not for the SGCN.

\end{itemize}

\subsection{Embeddings Comparison}
\begin{itemize}
    \item \textbf{Hypothesis:} The embeddings generated by SGCN and KASGCN are similar.
    \item \textbf{Results:} The embeddings for SGCN (Red) and KASGCN (Blue) are visualized using t-SNE for BitcoinAlpha, BitcoinOTC, and WikiElec in Figure~\ref{fig-tsne}. The average cosine similarity between the embeddings of SGCN and KASGCN for these three signed graphs is -0.02, 0.01, and 0.072, respectively.

\begin{figure*}[!ht]
    \centering
    \begin{minipage}{0.31\textwidth}
        \centering
        \includegraphics[width=\linewidth]{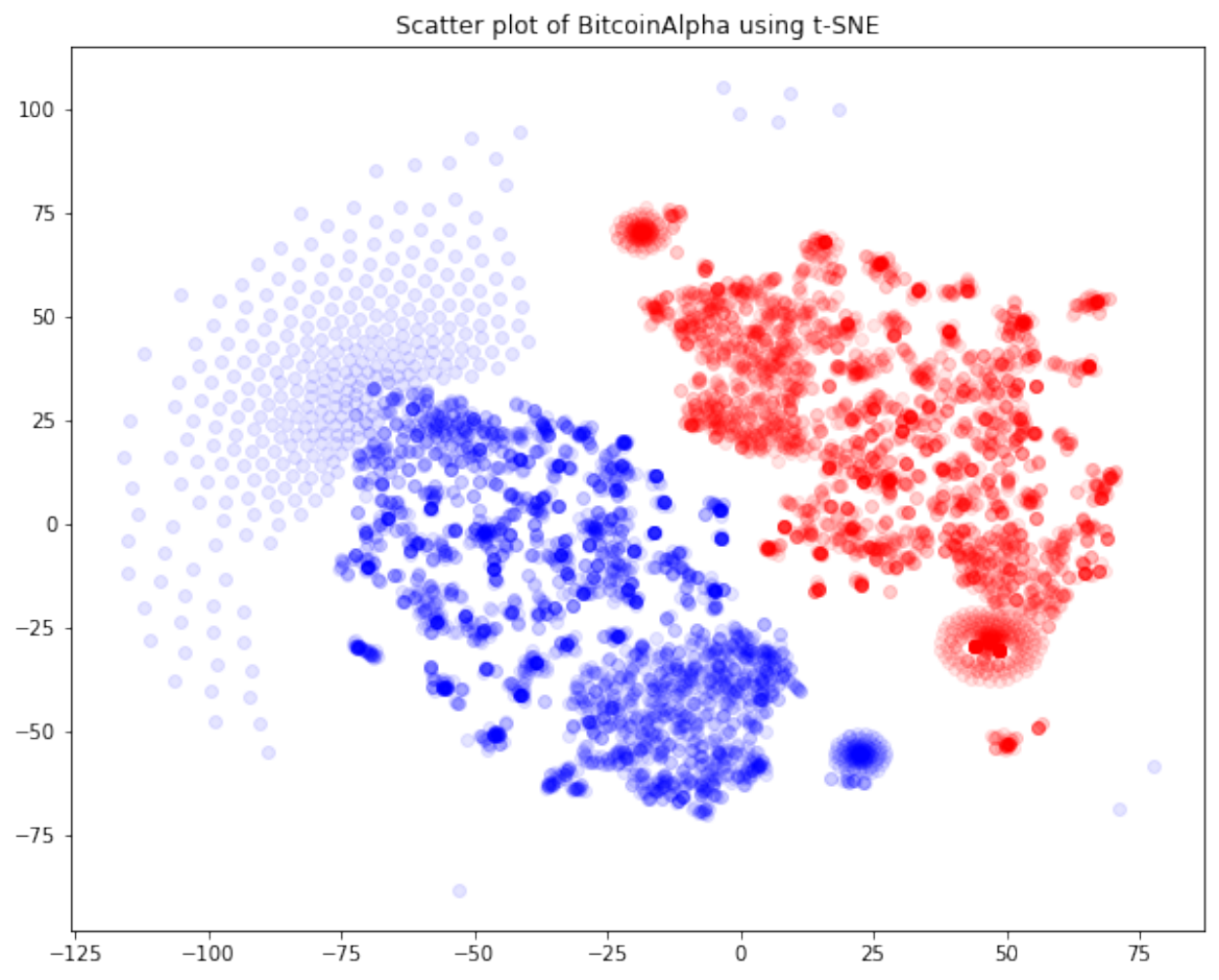}
    \end{minipage}
    \hfill
    \begin{minipage}{0.31\textwidth}
        \centering
        \includegraphics[width=\linewidth]{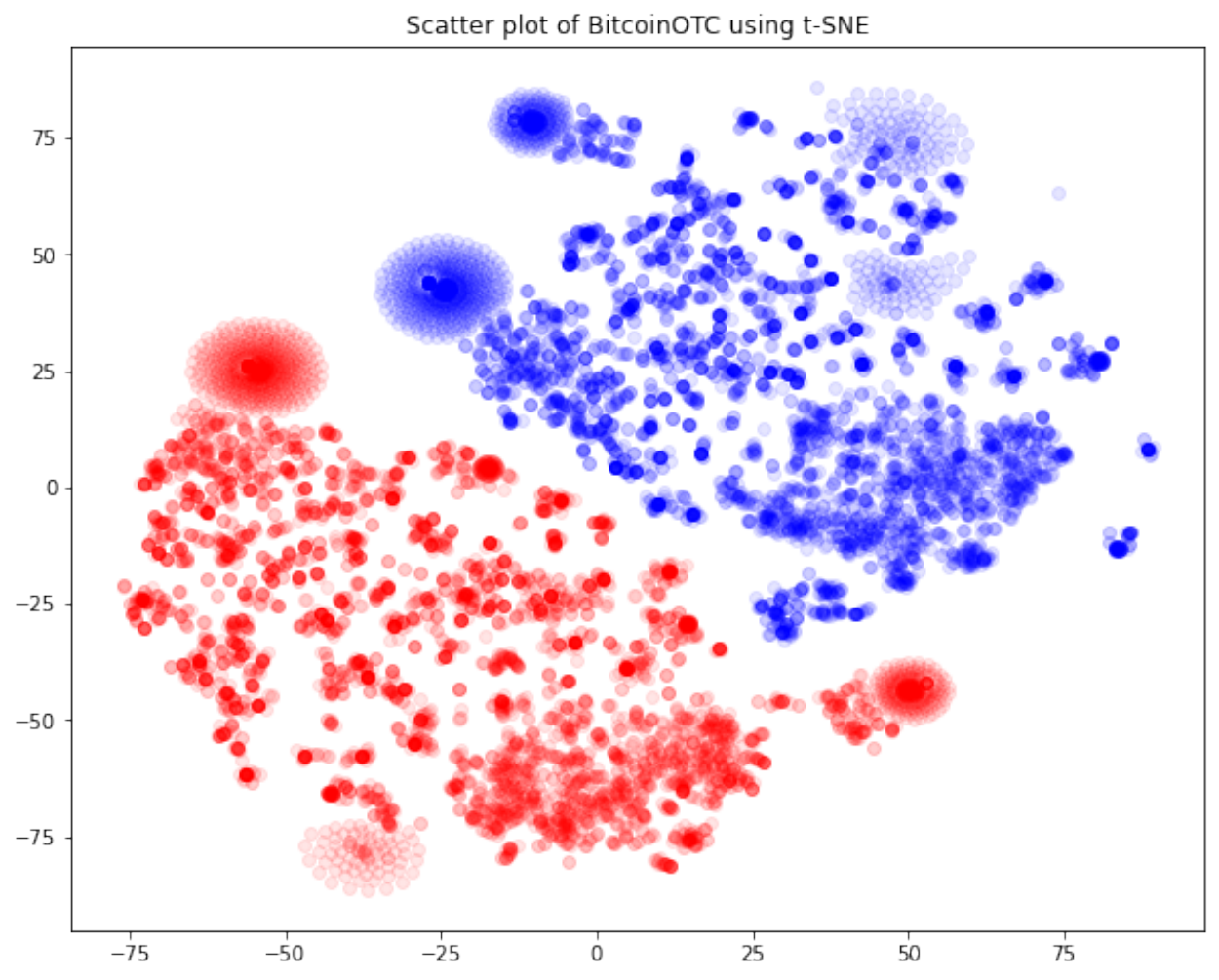}
    \end{minipage}
    \hfill
    \begin{minipage}{0.31\textwidth}
        \centering
        \includegraphics[width=\linewidth]{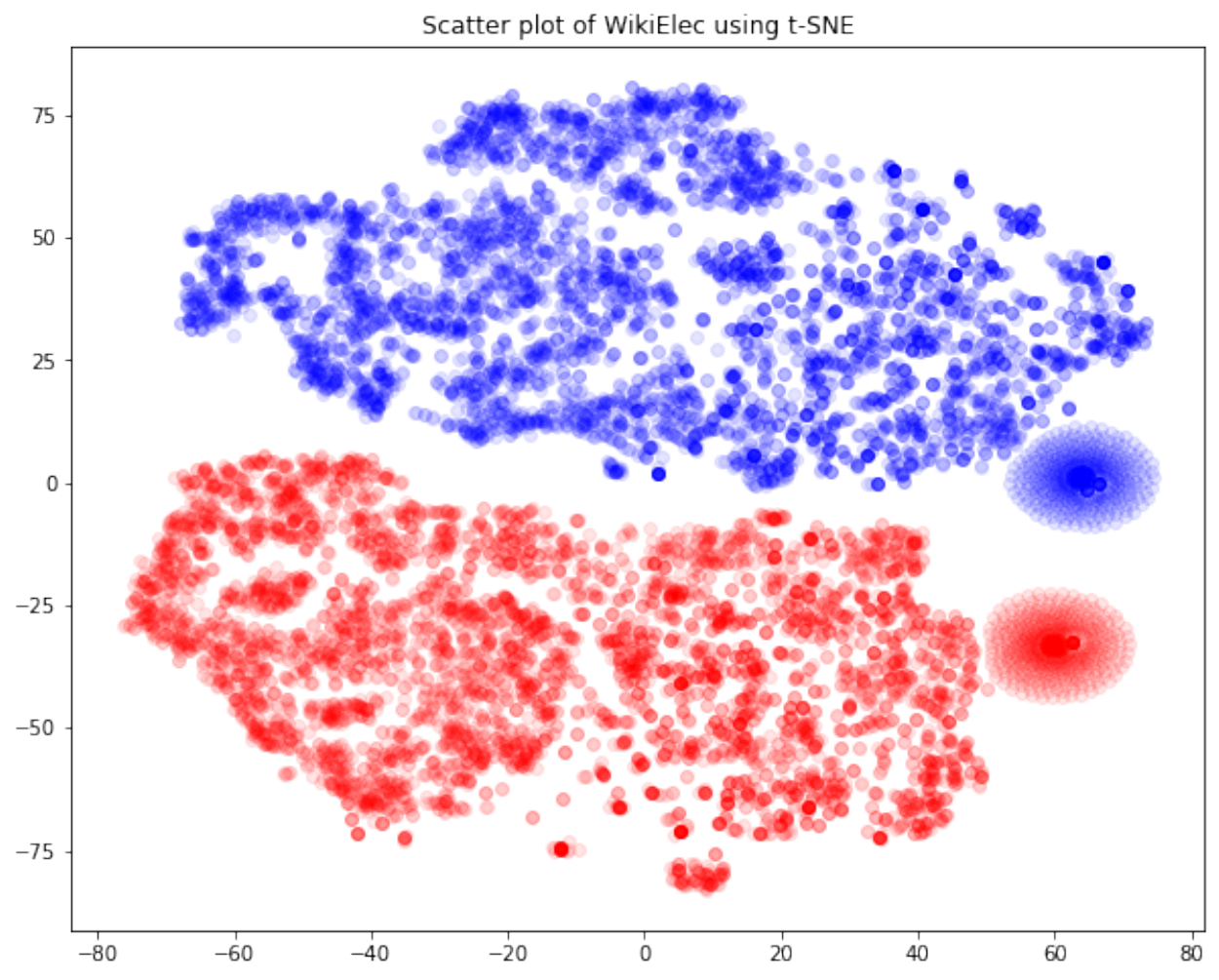}
    \end{minipage}
    \caption{t-SNE visualizations of the embeddings generated by SGCN (Red) and KASGCN (Blue) for different signed graphs.}
    \label{fig-tsne}
\end{figure*}

    \item \textbf{Observations:} Similarities are near zero, which indicates orthogonality. The t-SNE scatter plots show that the red and blue points are significantly dissimilar, highlighting that the embeddings do not match across the board.

\end{itemize}

\subsection{KAGCN Robustness}
\begin{itemize}
    \item \textbf{Hypothesis:} The embeddings generated by other SGCN integrated with other KAN variants outperform those generated by SGCN integrated with the original KAN layer in the community detection and link sign prediction tasks in the signed network.
    \item \textbf{Results:} The comparison of the community detection results between all the variants of the Kolmogorov-Arnold Layer is presented in Table~\ref{tab-comvar1}, Table~\ref{tab-comvar2}, and Table~\ref{tab-comvar3} for $K = 5$, $K = 10$, $K = 15$ respectively. Table~\ref{tab-linkvar} shows the link sign prediction results of the Kolmogorov-Arnold Layer variants where a Multinomial Logistic Regression is trained on the embeddings of both SGCN and KASGCN with random 0.2 test size splits over 10 runs.

        \begin{sidewaystable*}[!ht]
    \setlength{\tabcolsep}{2pt} 
    \centering
    \caption{Average $pos\_{in}$ and $neg\_{out}$ over 10 runs for KASGCN, FourierKASGCN, LaplaceKASGCN, and WaveletKASGCN node embeddings using Kmeans++ clustering ($K = 5$). Numbers in \textbf{bold} indicate that the clustering quality $Q$ is the highest. \label{tab-comvar1}}
\begin{tabular}{l|cc|cc|cc|cc}    \toprule
        \textbf{GNN} &  \multicolumn{2}{c|}{\textbf{KASGCN}} &  \multicolumn{2}{c|}{\textbf{FourierKASGCN}}& \multicolumn{2}{c|}{\textbf{LaplaceKASGCN}}& \multicolumn{2}{c}{\textbf{WaveletKASGCN}} \\ 
       \textbf{Quality} & pos & neg & pos & neg & pos & neg & pos & neg \\ \midrule
BitcoinAlpha & 0.4093 $\pm$ 0.011 & 0.659 $\pm$ 0.05 & 0.37 $\pm$ 0.0003 & 0.766 $\pm$ 0.0007 & \bf 0.462 $\pm$ 0.014 & \bf 0.716 $\pm$ 0.019 & 0.3879 $\pm$ 0.0016 & 0.786 $\pm$ 0.0011 \\ BitcoinOTC & 0.523 $\pm$ 0.00033 & 0.707 $\pm$ 0.001 & 0.515 $\pm$ 0.004 & 0.841 $\pm$ 0.0004 & 0.467 $\pm$ 0.019 & 0.721 $\pm$ 0.062 & \bf 0.511 $\pm$ 0.0003 & \bf 0.833 $\pm$ 0.0001 \\ 
WikiElec & 0.5221 $\pm$ 0.004 & 0.784 $\pm$ 0.0009 & \bf 0.546 $\pm$ 0.0021 & \bf 0.7844 $\pm$ 0.0014 & 0.418 $\pm$ 0.0071 & 0.725 $\pm$ 0.003 & 0.472 $\pm$ 0.0 & 0.781 $\pm$ 0.0 \\
WikiRFA & 0.391 $\pm$ 0.0010 & 0.795 $\pm$ 0.0006 & 0.425 $\pm$ 0.0003 & 0.807 $\pm$ 0.0003 & 0.409 $\pm$ 0.0062 & 0.708 $\pm$ 0.010 &\bf 0.422 $\pm$ 0.019 &\bf 0.853 $\pm$ 0.0016 \\ 
Chess & 0.409 $\pm$ 0.015 & 0.619 $\pm$ 0.034 & \bf 0.380 $\pm$ 0.091 & \bf 0.759 $\pm$ 0.017 & 0.551 $\pm$ 0.007 & 0.451 $\pm$ 0.0078 & 0.345 $\pm$ 0.00043 & 0.717 $\pm$ 0.00039 \\
Congress & 0.566 $\pm$ 0.029 & 0.9584 $\pm$ 0.017 & 0.554 $\pm$ 0.031 & 0.988 $\pm$ 0.0097 & 0.290 $\pm$ 0.054 & 0.647 $\pm$ 0.033 & \bf 0.648 $\pm$ 0.057 & \bf 0.977 $\pm$ 0.014 \\
PPI & \bf 0.385 $\pm$ 0.084 & \bf  0.87 $\pm$ 0.0109 & 0.445 $\pm$ 0.015 & 0.781 $\pm$ 0.0027 & 0.838 $\pm$ 0.061 & 0.223 $\pm$ 0.101 & 0.365 $\pm$ 0.0023 & 0.810 $\pm$ 0.0067 \\ \bottomrule
    \end{tabular}
\end{sidewaystable*}

        \begin{sidewaystable*}[!ht]
    \setlength{\tabcolsep}{2pt} 
    \centering
    \caption{Average $pos\_{in}$ and $neg\_{out}$ over 10 runs for KASGCN, FourierKASGCN, LaplaceKASGCN, and WaveletKASGCN node embeddings using Kmeans++ clustering ($K = 10$). Numbers in \textbf{bold} indicate that the clustering quality $Q$ is the highest. \label{tab-comvar2}}
\begin{tabular}{l|cc|cc|cc|cc}    \toprule
        \textbf{GNN} &  \multicolumn{2}{c|}{\textbf{KASGCN}} &  \multicolumn{2}{c|}{\textbf{FourierKASGCN}}& \multicolumn{2}{c|}{\textbf{LaplaceKASGCN}}& \multicolumn{2}{c}{\textbf{WaveletKASGCN}} \\ 
       \textbf{Quality} & pos & neg & pos & neg & pos & neg & pos & neg \\ \midrule
BitcoinAlpha & \bf 0.299 $\pm$ 0.006 &\bf 0.877 $\pm$ 0.012 & 0.253 $\pm$ 0.003 & 0.896 $\pm$ 0.0031 & 0.352 $\pm$ 0.019 & 0.770 $\pm$ 0.0087 & 0.281 $\pm$ 0.0077 & 0.876 $\pm$ 0.0052 \\ 
BitcoinOTC & 0.421 $\pm$ 0.005 & 0.864 $\pm$ 0.051 & 0.401 $\pm$ 0.017 & 0.922 $\pm$ 0.0086 & 0.370 $\pm$ 0.014 & 0.834 $\pm$ 0.015 & \bf 0.4100 $\pm$ 0.0029 & \bf 0.9246 $\pm$ 0.005 \\ 
WikiElec & 0.283 $\pm$ 0.020 & 0.909 $\pm$ 0.0022 & \bf 0.2741 $\pm$ 0.001 & \bf 0.920 $\pm$ 0.00085 & 0.267 $\pm$ 0.013 & 0.823 $\pm$ 0.0027 & 0.252 $\pm$ 0.0036 & 0.910 $\pm$ 0.0018 \\ 
WikiRFA & 0.270 $\pm$ 0.033 & 0.917 $\pm$ 0.04 & 0.250 $\pm$ 0.0033 & 0.930 $\pm$ 0.0012 & 0.400 $\pm$ 0.017 & 0.715 $\pm$ 0.031 &\bf 0.269 $\pm$ 0.0016 & \bf 0.924 $\pm$ 0.0076 \\ 
Chess & 0.22 $\pm$ 0.0054 & 0.888 $\pm$ 0.0022 & \bf 0.243 $\pm$ 0.0021 & \bf 0.873 $\pm$ 0.0014 & 0.419 $\pm$ 0.013 & 0.592 $\pm$ 0.012 & 0.225 $\pm$ 0.0048 & 0.864 $\pm$ 0.0068 \\ 
Congress & 0.472 $\pm$ 0.050 & 0.992 $\pm$ 0.0074 & 0.482 $\pm$ 0.037 & 1.0 $\pm$ 0.0 & 0.182 $\pm$ 0.016 & 0.845 $\pm$ 0.036 & \bf 0.521 $\pm$ 0.037 & \bf 0.995 $\pm$ 0.0066 \\
PPI & \bf 0.364 $\pm$ 0.037 & \bf 0.936 $\pm$ 0.0093 & 0.353 $\pm$ 0.0254 & 0.884 $\pm$ 0.0059 & 0.732 $\pm$ 0.022 & 0.387 $\pm$ 0.035 & 0.289 $\pm$ 0.0087 & 0.922 $\pm$ 0.0083 \\ \bottomrule
    \end{tabular}
\end{sidewaystable*}

        \begin{sidewaystable*}[!ht]
    \setlength{\tabcolsep}{2pt} 
    \centering
    \caption{Average $pos\_{in}$ and $neg\_{out}$ over 10 runs for KASGCN, FourierKASGCN, LaplaceKASGCN, and WaveletKASGCN node embeddings using Kmeans++ clustering ($K = 15$). Numbers in \textbf{bold} indicate that the clustering quality $Q$ is the highest. \label{tab-comvar3}}
\begin{tabular}{l|cc|cc|cc|cc}    \toprule
        \textbf{GNN} &  \multicolumn{2}{c|}{\textbf{KASGCN}} &  \multicolumn{2}{c|}{\textbf{FourierKASGCN}}& \multicolumn{2}{c|}{\textbf{LaplaceKASGCN}}& \multicolumn{2}{c}{\textbf{WaveletKASGCN}} \\
       \textbf{Quality} & pos & neg & pos & neg & pos & neg & pos & neg \\  \midrule
BitcoinAlpha &\bf 0.245 $\pm$ 0.0074 &\bf 0.921 $\pm$ 0.006 & 0.219 $\pm$ 0.0095 & 0.932 $\pm$ 0.0074 & 0.277 $\pm$ 0.024 & 0.837 $\pm$ 0.020 & 0.227 $\pm$ 0.0089 & 0.933 $\pm$ 0.0053 \\  
BitcoinOTC & \bf 0.380 $\pm$ 0.17 & \bf 0.93 $\pm$ 0.014 & 0.317 $\pm$ 0.012 & 0.965 $\pm$ 0.0041 & 0.316 $\pm$ 0.020 & 0.874 $\pm$ 0.011 & 0.337 $\pm$ 0.0136 & 0.954 $\pm$ 0.0041 \\ 
WikiElec & \bf 0.227 $\pm$ 0.0090 & \bf 0.942 $\pm$ 0.04 & 0.214 $\pm$ 0.0081 & 0.946 $\pm$ 0.0042 & 0.189 $\pm$ 0.018 & 0.866 $\pm$ 0.010 & 0.166 $\pm$ 0.0081 & 0.937 $\pm$ 0.0012 \\
WikiRFA & \bf 0.203 $\pm$ 0.0051 & \bf 0.9470 $\pm$ 0.0050 & 0.194 $\pm$ 0.0074 & 0.954 $\pm$ 0.0018 & 0.301 $\pm$ 0.0071 & 0.813 $\pm$ 0.0072 & 0.198 $\pm$ 0.0090 & 0.948 $\pm$ 0.0028 \\ 
Chess & 0.181 $\pm$ 0.0024 & 0.922 $\pm$ 0.0020 & \bf 0.191 $\pm$ 0.0035 & \bf 0.919 $\pm$ 0.0020 & 0.374 $\pm$ 0.018 & 0.631 $\pm$ 0.019 & 0.178 $\pm$ 0.0048 & 0.912 $\pm$ 0.0026 \\ 
Congress & 0.392 $\pm$ 0.034 & 0.996 $\pm$ 0.0048 & 0.4286 $\pm$ 0.029 & 0.999 $\pm$ 0.0029 & 0.149 $\pm$ 0.023 & 0.906 $\pm$ 0.029 & \bf 0.439 $\pm$ 0.0300 & \bf 1.0 $\pm$ 0.0 \\
PPI & \bf 0.309 $\pm$ 0.031 & \bf 0.959 $\pm$ 0.0049 & 0.307 $\pm$ 0.023 & 0.909 $\pm$ 0.006 & 0.684 $\pm$ 0.026 & 0.458 $\pm$ 0.0396 & 0.251 $\pm$ 0.014 & 0.941 $\pm$ 0.0020 \\ \bottomrule
    \end{tabular}
\end{sidewaystable*}

        \begin{sidewaystable*}[!ht]
    \setlength{\tabcolsep}{2pt} 
    \centering
    \caption{Average AUC and F1 score over 10 runs for KASGCN, FourierKASGCN, LaplaceKASGCN, and WaveletKASGCN node embeddings using Multinomial Logistic Regression with random 0.2 test size split. Numbers in \textbf{bold} indicate that the AUC/F1 metric is the highest. \label{tab-linkvar}}
\begin{tabular}{l|cc|cc|cc|cc}    \toprule
        \textbf{GNN} &  \multicolumn{2}{c|}{\textbf{KASGCN}} &  \multicolumn{2}{c}{\textbf{FourierKASGCN}}& \multicolumn{2}{c|}{\textbf{LaplaceKASGCN}}& \multicolumn{2}{c}{\textbf{WaveletKASGCN}} \\ 
       \textbf{Quality} & AUC & F1 & AUC & F1 & AUC & F1 & AUC & F1\\ \midrule
BitcoinAlpha & \bf 0.799 $\pm$ 0.022 & 0.690 $\pm$ 0.094 & 0.796 $\pm$ 0.018 & 0.656 $\pm$ 0.19 & 0.582 $\pm$ 0.039 & 0.095 $\pm$ 0.302 & 0.782 $\pm$ 0.020 & \bf 0.705 $\pm$ 0.090 \\ 
BitcoinOTC &\bf 0.859 $\pm$ 0.0075 & 0.761 $\pm$ 0.068 & 0.848 $\pm$ 0.0082 & \bf 0.785 $\pm$ 0.082 & 0.513 $\pm$ 0.046 & 0.0020 $\pm$ 0.0044 & 0.855 $\pm$ 0.0088 & 0.773 $\pm$ 0.065 \\ 
WikiElec &\bf 0.837 $\pm$ 0.0045 & 0.787 $\pm$ 0.032 & 0.819 $\pm$ 0.014 & 0.793 $\pm$ 0.046 & 0.488 $\pm$ 0.024 & 0.00030 $\pm$ 0.0009 & 0.828 $\pm$ 0.015 & \bf 0.800 $\pm$ 0.043 \\ 
WikiRFA & 0.830 $\pm$ 0.0037 & 0.759 $\pm$ 0.017 & 0.830 $\pm$ 0.0040 & \bf 0.782 $\pm$ 0.030 & 0.496 $\pm$ 0.02 & 0.0 $\pm$ 0.0 & \bf 0.836 $\pm$ 0.0046 & 0.765 $\pm$ 0.0338 \\ 
Chess & 0.548 $\pm$ 0.0064 & 0.625 $\pm$ 0.041 & 0.550 $\pm$ 0.0059 & \bf 0.635 $\pm$ 0.068 & 0.508 $\pm$ 0.012 & 0.018 $\pm$ 0.035 & \bf 0.552 $\pm$ 0.0049 & 0.610 $\pm$ 0.047 \\ 
Congress & 0.499 $\pm$ 0.046 & 0.602 $\pm$ 0.077 & 0.544 $\pm$ 0.059 & \bf 0.656 $\pm$ 0.074 & 0.493 $\pm$ 0.086 & 0.013 $\pm$ 0.042 &\bf 0.563 $\pm$ 0.083 & 0.590 $\pm$ 0.069 \\ 
PPI & 0.692 $\pm$ 0.015 & 0.654 $\pm$ 0.037 & 0.689 $\pm$ 0.019 & \bf 0.661 $\pm$ 0.071 & 0.513 $\pm$ 0.030 & 0.0 $\pm$ 0.0 & \bf 0.694 $\pm$ 0.016 & 0.641 $\pm$ 0.047 \\ \bottomrule
    \end{tabular}
\end{sidewaystable*}

    \item \textbf{Observations:} In the community detection task, identifying the best KASGCN variant is challenging, as performance appears to be influenced by both the K value and the signed graph. However, LaplaceKASGCN consistently underperforms across various graphs. For the link sign prediction task and across all datasets, LaplaceKASGCN also shows poor performance, which may not be suitable for this task. In contrast, FourierKASGCN and WaveletKASGCN demonstrate superior performance on most graphs, as evidenced by their higher F1 and AUC metrics.

\end{itemize}
\section{Conclusion and Future Work}
\label{sec-summ}
In this work, we investigated the integration of the Kolmogorov-Arnold Neural Network (KAN) within the Signed Graph Convolutional Network (SGCN) framework, aiming to evaluate its effectiveness in community detection and link sign prediction in signed networks. Our empirical results show that KAN-enhanced SGCNs perform comparably to traditional, unmodified SGCNs. However, the variability in performance, highlighted by relatively large standard deviations, suggests that the effectiveness of this approach is highly context-dependent, influenced by factors such as the specific characteristics of the signed graph and model parameters. Notably, the LaplaceKASGCN variant consistently performs poorly in signed network downstream tasks. The findings in this paper lay the groundwork for future exploration into the broader applicability of KANs in various domains. Specifically for signed graph networks, this research offers valuable insights for further refining the integration of KANs into graph-based learning models, paving the way for more sophisticated and effective analysis in the future.


\end{document}